\documentclass[conference]{IEEEtran}
\IEEEoverridecommandlockouts
\usepackage{cite}
\usepackage{amsmath,amssymb,amsfonts}
\usepackage{algorithmic}
\usepackage{graphicx}
\usepackage{textcomp}
\usepackage{xcolor}
\usepackage{svg}
\usepackage[hidelinks]{hyperref}

\def\BibTeX{{\rm B\kern-.05em{\sc i\kern-.025em b}\kern-.08em
    T\kern-.1667em\lower.7ex\hbox{E}\kern-.125emX}}
\begin{document}

\title{ME-CapsNet: A Multi-Enhanced Capsule Networks with Routing Mechanism\\
}

\author{\IEEEauthorblockN{Jerrin Bright\IEEEauthorrefmark{2},
Suryaprakash Rajkumar\IEEEauthorrefmark{2} and Arockia Selvakumar Arockia Doss\IEEEauthorrefmark{1} }
\IEEEauthorblockA{School of Mechanical Engineering,
Vellore Institute of Technology, Chennai, India\\
Email:jerriebright@gmail.com,
suryaprakash.rajkumar@gmail.com,
\IEEEauthorrefmark{1}arockiaselvakumar@vit.ac.in}
\IEEEauthorblockA{\IEEEauthorrefmark{2}These Authors Contributed Equally To This Work}}
\maketitle

\begin{abstract}
Convolutional Neural Networks need the construction of informative features, which are determined by channel-wise and spatial-wise information at the network's layers. In this research, we focus on bringing in a novel solution that uses sophisticated optimization for enhancing both the spatial and channel components inside each layer's receptive field. Capsule Networks were used to understand the spatial association between features in the feature map. Standalone capsule networks have shown good results on comparatively simple datasets than on complex datasets as a result of the inordinate amount of feature information. Thus, to tackle this issue, we have proposed ME-CapsNet by introducing deeper convolutional layers to extract important features before passing through modules of capsule layers strategically to improve the performance of the network significantly. The deeper convolutional layer includes blocks of Squeeze-Excitation networks which use a stochastic sampling approach for progressively reducing the spatial size thereby dynamically recalibrating the channels by reconstructing their interdependencies without much loss of important feature information. Extensive experimentation was done using commonly used datasets demonstrating the efficiency of the proposed ME-CapsNet, which clearly outperforms various research works by achieving higher accuracy with minimal model complexity in complex datasets.
\end{abstract}


\section{Introduction}
\label{sec:intro}

Convolutional Neural Network (CNN) is a cutting-edge image classification \cite{imageclass, imageclass2, imageclass3} approach that excels in detecting characteristics but struggles to grasp the spatial connection between them, such as their direction, perspective, and size. For scene understanding or any vision-based strategy, improving categorization accuracy is critical. In addition, transmitting information from one layer to the next (Routing) is less efficient in CNNs. This may be improved by identifying important characteristics using a variety of representations and building well-organized routing strategies for transferring data between layers. Investigation of many elements of network architecture (spatial, channel) is now a very prominent research subject in CNN, and it may greatly improve the network's understanding ability \cite{review}.

\begin{figure}
\centering
\includegraphics[width=8.8 cm]{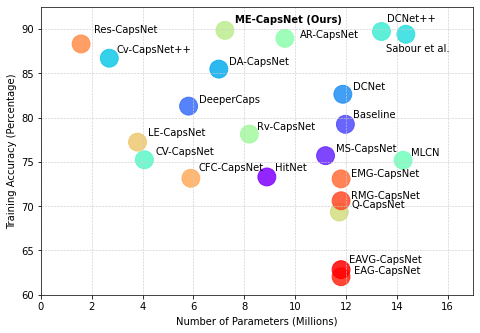}
\caption{Comparing various Capsule Network variations based on model complexity and test accuracy using CIFAR10 dataset.\label{fig1}}
\end{figure}   

Capsule Networks (CapsNet) where introduced by Sabour et al. \cite{sara2017} to solve the common issues with CNNs (rotational invariance and failure to capture spatial hierarchical information) by fetching most important feature information and spatial relationship efficiently. Instead of pooling operations, dynamic routing was adapted to prevent losing important features. Also, capsules where used instead of neurons thereby passing the spatial relationship from one layer to next efficiently. But, CapsNet tends to fail as a result of more feature information in complex datasets. Thus, adding of deeper convolutional layers to the CapsNet in a strategic way is one approach to solve this issue. 

Thus, we are proposing a novel network called Multi-Enhanced Capsule Networks (ME-CapsNet) which finds robust features, obtaining good channel and spatial relationships between layers. Squeeze-Excitation network or SENets \cite{Hu2018} with very little computation difference, models the interconnection of channels by adaptively recalibrating the channel features. Once the important features are learnt, the likelihood and the properties of each and every feature are estimated by passing it to the CapsNet module which inturn uses EM Routing to prevent feature losses. EM Routing \cite{emroute} is used to group capsules with the parent capsule using the closeness in the proximity of the corresponding estimated votes on the pose matrix. Also, reconstruction is included in the proposed work, unlike the originally proposed CapsNet with EM Routing.

\begin{figure*}[ht]
\centering
\includegraphics[width=18 cm]{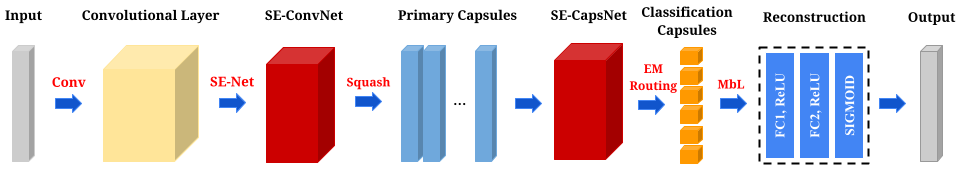}
\caption{Overall architecture of ME-CapsNet\label{fig2}}
\end{figure*}

Some contributions of this research work are mentioned here as follows: 

\begin{itemize}
\item	A novel network which multi enhances the features, that is, both spatial and channel-wise information at each layer with deep convolutional layers (SENets) and CapsNet;
\item  Improvements in the Squeeze phase of the originally proposed SENets by using Stochastic Spatial Sampling Pooling (S3P) to reduce feature information losses, computation overload and tranining time when compared to the originally used average pooling operation.
\item	Exploring the capabilities of the proposed system by fusing it with several state-of-the-art base models including ResNet, VGG, Inception and MobileNet and analyzing the performance of the proposed network by calibrating with different parameters.
\end{itemize}

The rest of the research paper is systematized as follows: Section \ref{works} consists of the related research works relating to various networks pertaining to SENets and CapsNets followed by Section \ref{proposal} detailing the proposed architecture along with its modules. Then in Section \ref{exp}, various experimentation is done testing by tweaking parameters and in Section \ref{res} the proposed architecture is tested with various state-of-the-art datasets and networks. Also, few predominantly efficient networks were handpicked and tested alongside our network. Finally, in Section \ref{conc} the proposed architecture is summarized with its advantages and is followed by future works premeditated to increase the robustness of the currently proposed system.

\section{Related Works} \label{works}

Since the introduction of deep convolutional networks \cite{deepcnn} significant improvement in the deep learning regime has been witnessed. Stacking of layers as a result of deep networks significantly improved the performance but also introduced vanishing gradient problems \cite{deeplearning}. Also, using operating layers like max pooling introduced loss of spatial information. CapsNet attempts to solve these issues by nesting layers instead of stacking them. These recent techniques in deep learning, added fuel to research in this field drastically.

CapsNet are considered a great leap into the neural networks considering its significance and scope of improvement over the traditional techniques. Some efficient research works on CapsNet are explained here. Mazzia et al. \cite{efficientnet} proposed a non-iterative routing technique that exploits self-attention mechanisms. Deliege et al. \cite{hitnet} introduced a counterpart for DigitCaps layer using a custom Hit-or-Miss layer. Huang et al. \cite{dacapsnet} proposed addition of two attention layers to the original CapsNet, to improve the overall performance of the network. 

SENets are predominantly used recently considering the significant improvement in performance with minimal amount of computation rise. A lot of customization to SENets have also been done, including Yue Cao et al. \cite{cao2019} performing a unification of NLnet and SENet resulting in a Global Context (GC) network. A lot of SENets applications has been established including some works by Qiu et al. \cite{qui2018}, Shunjun Wei et al. \cite{wei2020}, and Han et al. \cite{han2020} where fusion of SENets were done with various architectures of CNN for fish image classifications, PRI modulation recognition, sea ice image sensing respectively.

Yang et al. \cite{rscapsnet} proposes an approach to enhance the feature extraction capability of CapsNet using a modified version of ResNet called Res2Net and SENet in the base layers to enhance the performance of the CapsNet. This is the closest work in relavance to our work. We have strategically employed the SE blocks in between various layers and used a more powerful variant of pooling to improve the performance of the squeeze operation. 

\section{Methodology} \label{proposal}

In this section, the different components equipped by our work to complete the robust architecture is discussed extensively. Our architecture predominantly has three blocks to multi-enhance the feature relationship of CNN: the SE Primary Capsule (SE-PC) block, SE Classification Capsule (SE-CC) block and the Mask by Layer (MbL) block. That is, an input image in form of tensor (\emph{U}) is fed to our network, which first converts into a tensor upon transformation ($U_{tr}$). Then, $U_{tr}$ is sent through the the SE-PC block wherein Squeeze and Excitation operation is done which results in  $U_{tr}^o$ which has recalibrated weights based on channel dependence. $U_{tr}^o$ is then passed through the primary capsule layer. The Squeeze and Excitation operation is done three times continuously to get the improved calibrated weights for the corresponding channels. The Squeeze operation is done using S3P \cite{s3p} which will be explained in Section \ref{s3.1}. Then the output of the primary layer of the SE-PC is fed as input to the SE-CC wherein features are highlighted using SENets first and then is sent to the classification module. Then, we designed the MbL block inspired by the works for Huang et al. \cite{Huang} for reconstructing the output capsules from the probabilty vectors obtained from the SE-CC block. Advantages of including the reconstruction block include regularization and ability to regenerate data.

\subsection{Squeeze-Excitation Network} \label{s3.1}

SENets directly influence the performance of the network and its layers through adaptive adjusting of the weights obtained from each and every feature map in the channels of the convolutional block by adding supplementary parameters. This interfacing with weights radically increases the sensitivity to beneficial features in the map which will be exploited in succeeding transformations.

\begin{figure}[ht]
\centering
\includegraphics[width=7.5 cm]{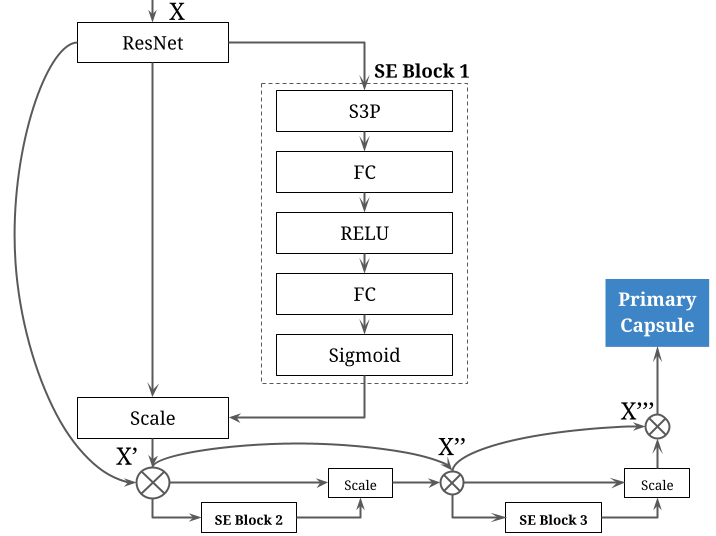}
\caption{Schema of the input (SE-ResNet) architecture for Capsule Network\label{fig3}}
\end{figure}  

Generally, in the squeeze phase, Global Average Pooling (GAP) or Maximum Pooling (GMP) techniques are equipped, each has its limitations and advantages. GAP (represented as $z_{avg}$ in equation \ref{eq1} results in smooth average pixels and doesn't preserve pixel information. GAP takes into consideration all the features, thus it can result in a biased feature map.

\begin{equation}
z_{avg}=\frac{1}{H \cdot W} \sum_{L=1}^{H} {\sum_{j=1}^{W} {u_c(i,j)}} \label{eq1}
\end{equation}

GMP (represented as $z_{max}$ in equation \ref{eq2}) extracts low-level features (edges, textures) better than GAP but captures a lot of noise and is independent of neighboring pixels. 

\begin{equation}
z_{max}=max_{i,j} \cdot {u_c(i,j)} \label{eq2}
\end{equation}

In this work, we will be using a pooling approach proposed by Zhai et al. \cite{s3p}. A two step process including GMP ($z_{max}$) and stochastic down sampling ($\mathcal{D}_g^s(.)$) is done on the feature map to learn better abstract representations in a larger receptive field. Considering it is stochastic, different feature map is obtained for the same training sample. This means it performs a virtual data augmentation \cite{data_aug}. Equation \ref{eq3} defines the S3P pooling, where \emph{s} represents strides and \emph{g} represents the grid size.

\begin{equation}
z_{s3p}= \mathcal{D}_g^s \cdot (z_{max}) \label{eq3}
\end{equation}

Now, the output of the squeeze phase ($z_mix$) is passed to the excitation phase for adaptive scaling of weights in the respective channel. Excitation is done using the Multi-Layer Perceptron (MLP) approach. That is, two fully connected layers are used with weights $W_1$ and $W_2$. The first fully connected layer is connected to ReLU activation \cite{relu} followed by another fully connected layer, which in turn is connected to the Sigmoid activation function. This can be represented as shown in equation \ref{eq4}. While training, $W_1$ and $W_2$ are learned. 

\begin{equation}
s= \sigma \cdot (W_2 \cdot \delta(W_1 \cdot z_{mix})) \label{eq4}
\end{equation}

In other words, channel-wise dependence is found inside the excitation phase by learning the weights, $W_1$ and $W_2$ using a compression ratio to enable efficient learning.

Figure \ref{fig3} shows the schema of the SENet with the Residual Network (ResNet) backbone leveraged in this work. In other words, the output of figure \ref{fig3} is the input of the primary capsule layer. A very similar architecture without ResNet, leveraging the output of capsule layers are used for the SENets equipped before the classification capsule layer. 

\subsection{Capsule Network}

Appropriate activation can be given if the likeliness of the feature and the feature information is passed via each and every neuron. Thus, these neurons are termed capsules and they output a vector (called an activity vector) instead of single feature information. Knowing the likeliness of a feature, we can easily extrapolate different feature variants in a capsule, thereby reducing the training data. The product of the pose matrix ($M_i$) represented by a gaussian distribution with the transformation ($W_{ij}$) is computed to calculate the vote ($v_{ij}$).

\begin{equation}
v_{ij}  = M_i \cdot W_{ij} \label{eq5}
\end{equation}

In equation \ref{eq5}, \emph{i} corresponds to the parent capsule and \emph{j} to the current capsule. The transformation is learned through back-propagation and a cost function.

\subsection{Routing Mechanism}

Transferring information from one layer to another is called Routing. ReLU is used as the routing mechanism in fully convolutional neural networks.

\begin{equation}
y_i = ReLU(\sum_j W_{ji} \cdot x_j +b_i)  \label{eq6}
\end{equation}

Instead of ReLU, a squashing function is used to convert small vector magnitudes to zero and big vector magnitudes to unit vectors. Grouping of capsules using the EM clustering technique is called EM Routing. Parent capsules are predicted in the pose matrix in terms of ‘votes’. It is based on proximity, so even when the viewpoints change, the transformation matrix remains constant. Assignment probabilities ($r_{ij}$) estimate the runtime connection with respect to the parent and other capsules.  The cost of all low-level capsules is shown in equation \ref{eq7}.

\begin{equation}
J_j^h= (ln(\sigma_j^h)+k) \sum_i r_{ij} \label{eq7}
\end{equation}

Activation ($A_j$) in a particular capsule can be estimated using the inverse temperature parameter ($\lambda$) and parameters iteratively estimated using EM Routing by equation \ref{eq8}.

\begin{equation}
A_j= sigmoid (\lambda(b_j - \sum_h J_j^h)) \label{eq8}
\end{equation}

The data points fit into a gaussian model and E and M steps are triggered alternatively iterating for ‘\emph{t}’ steps. The Gaussian model is updated using the M step based on the $r_{ij}$ initialized (distributed uniformly) which will then be reshaped to form the pose matrix of the parent capsule. The $r_{ij}$ is recomputed using the E-step for every data point.

Upon the completion of EM Routing, the capsules with best activation is selected to make the appropriate predictions using the MLP approach used in the excitation phase of the SE Blocks of ME-CapsNet.

\subsection{Reconstruction}

The output of the classification capsule is used for reconstruction of the input image by passing through a decoder. The output is a vector which consists the probability vector squashed to get values between 0 to 1. The idea behind the used non-linear squashing function is to compress small vectors to zero and long vectors to one. The decoder network consists of two fully connected layers with ReLU activation and a sigmoid function. Mean Square Error (MSE) is used for estimating the reconstruction loss between true ($y$) and predicted ($y^0$) labels as shown in equation \ref{eq:mse}.

\begin{equation}
    L(y, y^0) = \frac{1}{N} \sum^{N}_{i=0}(y - y^0) \label{eq:mse}
\end{equation}

\begin{figure*}
  \centering
  \begin{tabular}{ c @{\hspace{20pt}} c @{\hspace{20pt}} c @{\hspace{20pt}} c }

    \includegraphics[width=.42\columnwidth]{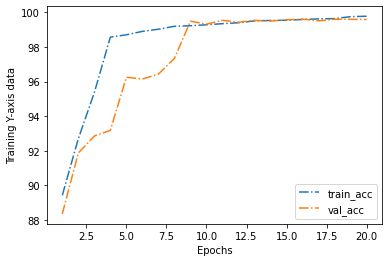} &
    \includegraphics[width=.42\columnwidth]{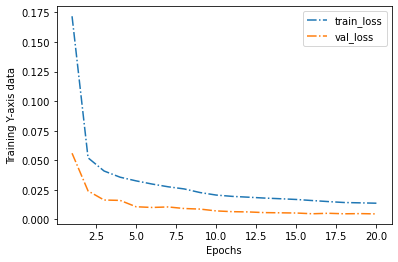} &
    \includegraphics[width=.42\columnwidth]{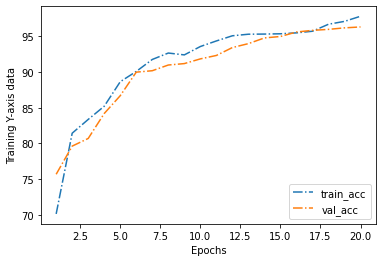} &
    \includegraphics[width=.42\columnwidth]{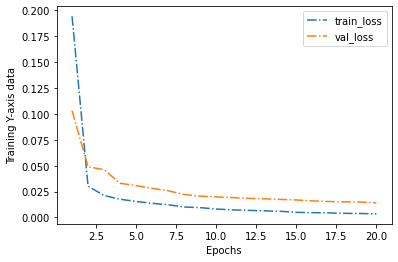}\\
    \small (a) MNIST Training Results & \small (b) MNIST Loss Results & \small (c) FMNIST Training Results & \small (d) FMNIST Loss Results \\
    
    \includegraphics[width=.42\columnwidth]{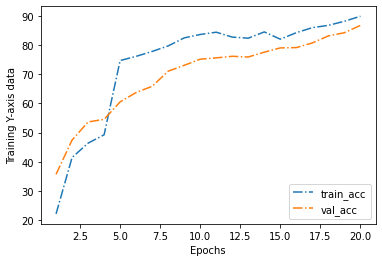} &
    \includegraphics[width=.42\columnwidth]{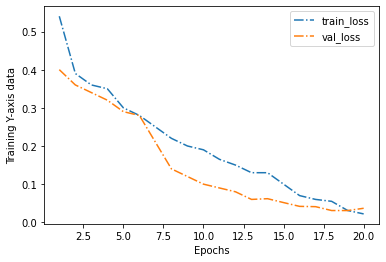} &
    \includegraphics[width=.42\columnwidth]{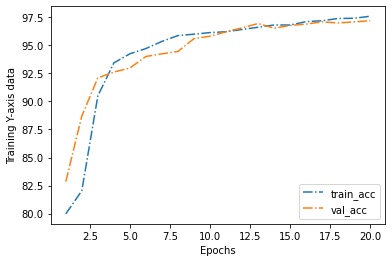} &
    \includegraphics[width=.42\columnwidth]{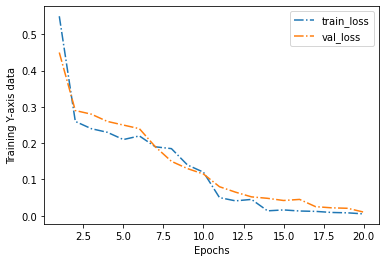}\\    
    \small (e) CIFAR10 Training Results & \small (f) CIFAR10 Loss Results & \small (g) KMNIST Training Results & \small (h) KMNIST Loss Results\\
      
  \end{tabular}

  \medskip

  \caption{Training and Validation results of accuracy and overall loss using various state-of-the-art datasets (MNIST, FashionMNIST- represented as FMNIST, CIFAR10, KMNIST)} \label{fig4}
\end{figure*}

\section{Experimentation} \label{exp}

In this section, experiments are steered from the architecture proposed in the previous section using MNIST, FashionMNIST and CIFAR-10 datasets, and the performance is evaluated with respect to other networks, previous effective research works and parameter tuning. 

Implementation of ME-CapsNet is done using several tuned parameters for efficient training and some important parameters are mentioned below. We have built ME-CapsNet using Keras and tensorflow framework. Tensorboard visualization toolkit was used for analyzing the performance of the proposed network. The network was compiled using the NVIDIA GTX 1650 GPU. The dataset was publicly available for the training of the model. For the training procedure, 20 epochs were done uniformly for all datasets, a batch size of 128, Adam optimizer \cite{Adam}, squashing constant of 0.5 and decay factor of 0.9.

The section has been split into three based on the target calibration. In section \ref{ss4.1}, the parameters of SENet is tweaked and the resulting top-1\% and top-10\% error rates are obtained. This is followed by section \ref{ss4.2}, where the capsule parameters are calibrated using various dimensions and quantity. 

\subsection{Capsule parameters} \label{ss4.2}

The number of capsules affects the overall accuracy and performance of the proposed ME-CapsNet. In table \ref{tab1}, different capsule dimensions are selected and the influence of capsule dimension on the overall accuracy of the network was tested. We have got the best accuracy when testing ME-CapsNet with dimension of capsule set to be 8. This testing has been done using CIFAR10 dataset.

\begin{table}[htbp]
    \centering
        \caption{Capsule dimension\label{tab1}}
        \resizebox{45mm}{!}{\begin{tabular}{|c|c|}
        \hline
            Dimension & Accuracy \\ 
            \hline
            8 & \textbf{89.84\%} \\ 
            \hline
            12 & 89.12\% \\ 
            \hline
            16 & 88.57\% \\
            \hline
        \end{tabular}}
\end{table}

In table \ref{tab2}, we tested our architecture with three variations in the total number of capsules and the respective accuracy has been logged corresponding to it.Best results where obtained when ME-CapsNet was tested with 5 capsules. This testing has been done using CIFAR10 dataset and the dimension of capsule set to be 8.

\begin{table}[htbp]
    \centering
        \centering
        \caption{No. of capsules\label{tab2}}
        \resizebox{45mm}{!}{\begin{tabular}{|c|c|}
        \hline
            Number & Accuracy \\ 
            \hline
            5 & 88.79\% \\ 
            \hline
            8 & \textbf{89.84\%} \\ 
            \hline
            10 & 86.44\% \\ 
            \hline
        \end{tabular}}
\end{table}

\subsection{SE Network parameters} \label{ss4.1}

As discussed in previous sections, SENets are split into Squeeze and Excitation operation. The squeeze operation consists of a pooling technique which includes max pooling, average pooling and the proposed S3P. All the pooling techniques are extensively studied in section \ref{s3.1}. Similarly, excitation operation is predominantly carried by the activation function like Sigmoid, ReLU, LeakyReLU and TanH. 

\begin{table}[htbp]
\centering
    \caption{Comparison with SE parameters.\label{tab3}}
    \resizebox{85mm}{!}{\begin{tabular}{|c|c|c|}
    \hline
        & top-1\% & top -5\% \\ 
        \hline
        Squeeze - Max & 22.57 & 6.09 \\ 
        \hline
        Squeeze - Avg & 22.28 & 6.03 \\ 
        \hline
        \textbf{Squeeze - S3P (Ours)} & \textbf{21.78} & \textbf{5.83} \\ 
        \hline
        Excitation - Sigmoid & 22.28 & 6.03 \\ 
        \hline
        Excitation - ReLU & 23.47 & 6.98 \\ 
        \hline
        Excitation - LeakyReLU & 23.22 & 6.91 \\ 
        \hline
        Excitation - TanH & 23.00 & 6.38 \\ 
        \hline
    \end{tabular}}
\end{table}

Sigmoid is an activation function that adds non-linearity and ranges between 0 to 1. Similarly, TanH is an activation function very similar to Sigmoid’s properties, but ranges from -1 to 1. ReLU is a linear function which outputs the input if positive and 0 if negative. Meanwhile, LeakyReLU works similar to ReLU for positive inputs, but outputs a very small positive value when the input tends to be negative.

\section{Results} \label{res}

In this section, ME-CapsNet equipped with the best parameters obtained from the experimentation done in Section \ref{exp} is compared alongside different backbone state-of-the-art models (Section \ref{ss5.1}) and novel research works pertaining to CapsNet (Section \ref{ss5.3}). Also details on the used datasets used for testing the propsed architecture is explained in \ref{dataset}.

\subsection{Datasets} \label{dataset}

Some commonly used datasets used to experiment the performance of ME-CapsNet include MNIST \cite{mnist}, FashionMNIST \cite{fashionmnist}, CIFAR10 \cite{cifar10} and KMNIST \cite{kmnist}. The performance of ME-CapsNet when tested alongside these datasets can be visualized in \ref{fig4}

MNIST and FashionMNIST both has 10 classes with 70,000 images of size $28\times28\times1$.CIFAR10 has 10 classes with 60,000 images of size $32\times32\times3$. KMNIST or Kuzushiji-MNIST dataset has 10 classes is a drop-in replacement of MNIST and FashionMNIST, consisting of 70,000 images of size $28\times28\times1$ just like those datasets. For MNIST, KMNIST and FashionMNIST datasets 50,000 images where used for training and 10,000 images for testing. For CIFAR10 dataset, 60,000 images where used for training and 10,000 for testing.

\subsection{Comparing various backbones} \label{ss5.1}

In this subsection, various backbone architectures including residual network with various levels of layers, inception network and VGG network is fused as input layer for ME-CapsNet and the respective test accuracy is logged and compared systematically for four different datasets. 

\begin{table}[htbp]
\centering
    \caption{Comparison of various backbone architectures\label{tab4}}
    \resizebox{85mm}{!}{\begin{tabular}{|c|c|c|c|c|c|}
    \hline
        & MNIST & FashionMNIST & CIFAR10 & KMNIST\\ 
        \hline
        ResNet-50 & 99.57 & 94.85 & 88.87 & 96.75\\ 
        \hline
        ResNet-101 & \textbf{99.83} & \textbf{95.63} & 89.37 & \textbf{97.84} \\ 
        \hline
        ResNet-152 & 99.78 & 95.91 & \textbf{89.84} & 97.11\\
        \hline
        VGG-16 \cite{karen2014} & 98.54 & 95.11 & 87.72 & 96.49\\ 
        \hline
        Inception \cite{Christian2014} & 98.63 & 95.38 & 88.98 & 97.34\\ 
        \hline
    \end{tabular}}
\end{table}

\subsection{Comparing architectures} \label{ss5.3}

In this subsection, some state-of-the-art research works aligning to CapsNet is studied in terms of accuracy, number of parameters and the ability to reconstruct. Table \ref{tab6} clearly shows our network, ME-CapsNet outperforms the other research works significantly with comparatively decent number of parameters using CIFAR10 dataset.  

\begin{table}[htbp]
    \centering
    \caption{Comparison with various proposed network.\label{tab6}}
    \resizebox{85mm}{!}{\begin{tabular}{|c|c|c|c|}
    \hline
        & Accuracy & \# of Params & Recon. \\ 
        \hline
            HitNet \cite{hitnet} & 73.30\% & 8.89M & Yes \\
            \hline
            MS-CapsNet \cite{mscapsnet} & 75.70\% & 11.20M & Yes \\
            \hline
            CapsNet Baseline & 79.24\% & 11.98M & No \\
            \hline
            DeeperCaps \cite{deepercaps} & 81.29\% & 5.81M & Yes \\
            \hline
            DCNet \cite{dcnet} & 82.63\% & 11.88M & Yes \\
            \hline
            DA-CapsNet \cite{dacapsnet} & 85.47\% & 7M & Yes \\
            \hline
            Cv-CapsNet++ \cite{cvcapsnet} & 86.70\% & 2.69M & No \\
            \hline
            AR-CapsNet \cite{arcapsnet} & 88.94\% & 9.60M & Yes \\
            \hline
            Sabour et. al. \cite{sara2017} & 89.40\% & 14.36M & Yes \\
            \hline
            DCNet++ \cite{dcnet} & 89.71\% & 13.4M & Yes \\
            \hline
            \textbf{ME-CapsNet (Ours)} & \textbf{89.84\%} & 7.24M & \textbf{Yes} \\ 
        \hline
    \end{tabular}}
\end{table}

ME-CapsNet easily outperforms the research works on CapsNet in CIFAR10 datasets, thus demonstrating its ability to capture important features and route without feature loss. Also, 1.23\% better performance in terms of accuracy was estimated using our model with FashionMNIST dataset. Though the difference in accuracy between the proposed approach and the previous best approach is only 0.13\%, computationally our approach tends to be very computationally cheaper. This is because of the less computational increase when using deep convolutional layers via SENets in between the CapsNet layers strategically with well-defined parameters.

\section{Conclusion and Future Work} \label{conc}

In this research, we have proposed a novel deep learning architecture that simultaneously enhances the spatial and channel-wise relationship of features with minimum computation. This multi-enhancement based network has been coined ME-CapsNet which uses a stochastic pooling approach to strategically improve the performance of the SENets used in CapsNet. The originally proposed CapsNet was used as the base network and SENets have been fused strategically before and in between capsule layers. The squeeze operation proposed in the original SENets where altered in this work using S3P, a stochastic sampling-based pooling approach which in turn significantly improved the performance of the SENets used in the proposed ME-CapsNet. The architecture was evaluated with state-of-the-art datasets and comparatively better results in terms of accuracy and computational cost were obtained upon testing with the proposed model.

Future study will focus on evaluating ME-CapsNet on broad, complex datasets. Though the aim of the originally proposed CapsNet was to prevent the use of pooling operations to ward off the loss of information, ME-CapsNet like many other networks uses the advantages of pooling operators to enhance the performance of CapsNets especially for complex datasets. One right research direction to proceed from here would be to look for alternatives of pooling operators for increasing the performance especially for complex datasets. Also, adding attention to routing mechanism can be added to our network to improve its performance. Computational time can be further reduced by replacing the fully connected layers with capsule layers. 

\bibliographystyle{ieeetr}
\bibliography{ref}

\end{document}